# Stacking classifiers for anti-spam filtering of e-mail

Georgios Sakkis♦, Ion Androutsopoulos♣, Georgios Paliouras♣, Vangelis Karkaletsis♣,
Constantine D. Spyropoulos♣, and Panagiotis Stamatopoulos♦

♦Department of Informatics
University of Athens
TYPA Buildings, Panepistimiopolis
GR-157 71 Athens, Greece
e-mail: {stud0926,
T.Stamatopoulos}@di.uoa.gr

♣Software and Knowledge Engineering Laboratory
Institute of Informatics and Telecommunications
National Centre for Scientific Research
"Demokritos"
GR-153 10 Ag. Paraskevi, Athens, Greece
e-mail: {ionandr, paliourg, vangelis, costass}@iit.demokritos.gr

## Abstract

We evaluate empirically a scheme for combining classifiers, known as stacked generalization, in the context of anti-spam filtering, a novel cost-sensitive application of text categorization. Unsolicited commercial e-mail, or "spam", floods mailboxes, causing frustration, wasting bandwidth, and exposing minors to unsuitable content. Using a public corpus, we show that stacking can improve the efficiency of automatically induced anti-spam filters, and that such filters can be used in real-life applications.

## Introduction

This paper presents an empirical evaluation of *stacked generalization*, a scheme for combining automatically induced classifiers, in the context of *anti-spam filtering*, a novel cost-sensitive application of text categorization.

The increasing popularity and low cost of e-mail have intrigued direct marketers to flood the mailboxes of thousands of users with unsolicited messages, advertising anything, from vacations to get-rich schemes. These messages, known as *spam* or more formally *Unsolicited Commercial E-mail*, are extremely annoying, as they clutter mailboxes, prolong dial-up connections, and often expose minors to unsuitable content (Cranor & Lamacchia, 1998).

Legal and simplistic technical counter-measures, like blacklists and keyword-based filters, have had a very limited effect so far.[1] The success of machine learning techniques in text categorization (Sebastiani, 2001) has recently led to alternative, learning-based approaches (Sahami, et al. 1998; Pantel & Lin, 1998; Drucker, et al. 1999). A classifier capable of distinguishing between spam and non-spam, hereafter *legitimate*, messages is induced from a manually categorized learning collection of messages, and is then used to identify incoming spam e-mail. Initial results have been promising, and experiments are becoming more systematic, by exploiting recently introduced benchmark corpora, and cost-sensitive evaluation measures (Gomez Hidalgo, et al. 2000; Androutsopoulos, et al. 2000a, b, c).

Stacked generalization (Wolpert, 1992), or *stacking*, is an approach for constructing classifier ensembles. A *classifier ensemble*, or *committee*, is a set of classifiers whose individual decisions are combined in some way to classify new instances (Dietterich, 1997). Stacking combines multiple classifiers to induce a higher-level classifier with improved performance. The latter can be thought of as the *president* of a committee with the ground-level classifiers as *members*. Each unseen incoming message is first given to the members; the president then decides on the category of the message by considering the opinions of the members and the message itself. Ground-level classifiers often make different classification errors. Hence, a president that has successfully learned when to trust each of the members can improve overall performance.

---

[1] Consult www.cauce.org, spam.abuse.net, and www.junkemail.org.

We have experimented with two ground-level classifiers for which results on a public benchmark corpus are available: a Naïve Bayes classifier (Androutsopoulos, et al. 2000a, c) and a memory-based classifier (Androutsopoulos, et al. 2000b; Sakkis, et al. 2001). Using a third, memory-based classifier as president, we investigated two versions of stacking and two different cost-sensitive scenarios. Overall, our results indicate that stacking improves the performance of the ground-level classifiers, and that the performance of the resulting anti-spam filter is acceptable for real-life applications.

Section 1 below presents the benchmark corpus and the preprocessing of the messages; section 2 introduces cost-sensitive evaluation measures; section 3 provides details on the stacking approaches that were explored; section 4 discusses the learning algorithms that were employed and the motivation for selecting them; section 5 presents our experimental results followed by conclusions.

# 1 Benchmark corpus and preprocessing

Text categorization has benefited from public benchmark corpora. Producing such corpora for anti-spam filtering is not straightforward, since user mailboxes cannot be made public without considering privacy issues. A useful public approximation of a user's mailbox, however, can be constructed by mixing spam messages with messages extracted from spam-free public archives of mailing lists. The corpus that we used, Ling-Spam, follows this approach (Androutsopoulos, et al. 2000a, b; Sakkis, et al. 2001). It is a mixture of spam messages and messages sent via the Linguist, a moderated list about the science and profession of linguistics. The corpus consists of 2412 Linguist messages and 481 spam messages.

Spam messages constitute 16.6% of Ling-Spam, close to the rates reported by Cranor and LaMacchia (1998), and Sahami et al. (1998).

Although the Linguist messages are more topic-specific than most users' e-mail, they are less standardized than one might expect. For example, they contain job postings, software availability announcements and even flame-like responses. Moreover, recent experiments with an encoded user mailbox and a Naïve Bayes (NB) classifier (Androutsopoulos, et al. 2000c) yielded results similar to those obtained with Ling-Spam (Androutsopoulos, et al. 2000a). Therefore, experimentation with Ling-Spam can provide useful indicative results, at least in a preliminary stage. Furthermore, experiments with Ling-Spam can be seen as studies of anti-spam filtering of open unmoderated lists.

Each message of Ling-Spam was converted into a vector $\vec{x} = \langle x_1, x_2, x_3, \ldots, x_n \rangle$, where $x_1, \ldots, x_n$ are the values of attributes $X_1, \ldots, X_n$. Each attribute shows if a particular word (e.g. "adult") occurs in the message. All attributes are binary: $X_i = 1$ if the word is present; otherwise $X_i = 0$. To avoid treating forms of the same word as different attributes, a lemmatizer was applied, converting each word to its base form.

To reduce the dimensionality, attribute selection was performed. First, words occurring in less than 4 messages were discarded. Then, the Information Gain (*IG*) of each candidate attribute $X$ was computed:

$$IG(X,C) = \sum_{x \in \{0,1\}, c \in \{spam, legit\}} P(x,c) \cdot \log \frac{P(x,c)}{P(x) \cdot P(c)}$$

The attributes with the *m* highest *IG*-scores were selected, with *m* corresponding to the best configurations of the ground classifiers that have been reported for Ling-Spam (Androutsopoulos, et al. 2000a; Sakkis, et al. 2001); see Section 4.

# 2 Evaluation measures

Blocking a legitimate message is generally more severe an error than accepting a spam message. Let $L \to S$ and $S \to L$ denote the two error types, respectively, and let us assume that $L \to S$ is λ times as costly as $S \to L$.

Previous research has considered three cost scenarios, where λ = 1, 9, or 999

(Androutsopoulos, et al. 2000a, b, c; Sakkis, et al. 2001). In the scenario where $\lambda = 999$, blocked messages are deleted immediately. $L \rightarrow S$ is taken to be 999 times as costly as $S \rightarrow L$, since most users would consider losing a legitimate message unacceptable. In the scenario where $\lambda = 9$, blocked messages are returned to their senders with a request to resend them to an unfiltered address. In this case, $L \rightarrow S$ is penalized more than $S \rightarrow L$, to account for the fact that recovering from a blocked legitimate message is more costly (counting the sender's extra work) than recovering from a spam message that passed the filter (deleting it manually). In the third scenario, where $\lambda = 1$, blocked messages are simply flagged as possibly spam. Hence, $L \rightarrow S$ is no more costly than $S \rightarrow L$. Previous experiments indicate that the Naïve Bayes ground-classifier is unstable when $\lambda = 999$ (Androutsopoulos, et al. 2000a). Hence, we have considered only the cases where $\lambda = 1$ or 9.

Let $W_L(\vec{x})$ and $W_S(\vec{x})$ be the confidence of a classifier (member or president) that message $\vec{x}$ is legitimate and spam, respectively. The classifier classifies $\vec{x}$ as spam iff:

$$\frac{W_S(\vec{x})}{W_L(\vec{x})} > \lambda$$

If $W_L(\vec{x})$ and $W_S(\vec{x})$ are accurate estimates of $P(legit | \vec{x})$ and $P(spam | \vec{x})$, respectively, the criterion above achieves optimal results (Duda & Hart, 1973).

To measure the performance of a filter, *weighted accuracy* (*WAcc*) and its complementary *weighted error rate* (*WErr* = 1 − *WAcc*) are used (Androutsopoulos, et al. 2000a, b, c; Sakkis, et al. 2001):

$$WAcc = \frac{\lambda \cdot N_{L \rightarrow L} + N_{S \rightarrow S}}{\lambda \cdot N_L + N_S}$$

where $N_{Y \rightarrow Z}$ is the number of messages in category $Y$ that the filter classified as $Z$, $N_L = N_{L \rightarrow L} + N_{L \rightarrow S}$, $N_S = N_{S \rightarrow S} + N_{S \rightarrow L}$. That is, when a legitimate message is blocked, this counts as $\lambda$ errors; and when it passes the filter, this counts as $\lambda$ successes.

We consider the case where no filter is present as our baseline: legitimate messages are never blocked, and spam messages always pass. The weighted accuracy of the baseline is:

$$WAcc^b = \frac{\lambda \cdot N_L}{\lambda \cdot N_L + N_S}$$

The *total cost ratio* (*TCR*) compares the performance of a filter to the baseline:

$$TCR = \frac{WErr^b}{WErr} = \frac{N_S}{\lambda \cdot N_{L \rightarrow S} + N_{S \rightarrow L}}$$

Greater *TCR* values indicate better performance. For *TCR* < 1, not using the filter is better.

Our evaluation measures also include *spam recall* (*SR*) and *spam precision* (*SP*):

$$SR = \frac{N_{S \rightarrow S}}{N_{S \rightarrow S} + N_{S \rightarrow L}}$$

$$SP = \frac{N_{S \rightarrow S}}{N_{S \rightarrow S} + N_{L \rightarrow S}}$$

*SR* measures the percentage of spam messages that the filter blocks (intuitively, its effectiveness), while *SP* measures how many blocked messages are indeed spam (its safety). Despite their intuitiveness, comparing different filter configurations using *SR* and *SP* is difficult: each configuration yields a pair of *SR* and *SP* results; and without a single combining measure, like *TCR*, that incorporates the notion of cost, it is difficult to decide which pair is better.

In all the experiments, *stratified 10-fold cross-validation* was used. That is, Ling-Spam was partitioned into 10 equally populated parts, maintaining the original spam-legitimate ratio. Each experiment was repeated 10 times, each time reserving a different part $S_j$ ($j = 1, ..., 10$) for testing, and using the remaining 9 parts as the training set $L_j$.

## 3 Stacking

In the first version of stacking that we explored (Wolpert, 1992), which we call *cross-validation stacking*, the training set of the president was prepared using a second-level 3-fold cross-validation. Each training set $L_j$ was further partitioned into three equally populated parts, and the training set of the president was prepared in three steps. At each step, a different part $LS_i$ ($i = 1, 2, 3$) of $L_j$ was reserved, and the members were trained on the union $LL_i$ of the other two parts. Each $\vec{x} = \langle x_1, \ldots, x_m \rangle$ of $LS_i$ was enhanced with the members' confidence $W_S^1(\vec{x})$ and $W_S^2(\vec{x})$ that $\vec{x}$ is spam, yielding an enhanced $LS_i'$ with vectors $\vec{x}' = \langle x_1, \ldots, x_m, W_S^1(\vec{x}), W_S^2(\vec{x}) \rangle$. At the end of the 3-fold cross-validation, the president was trained on $L_j' = LS_1' \cup LS_2' \cup LS_3'$. It was then tested on $S_j$, after retraining the members on the entire $L_j$ and enhancing the vectors of $S_j$ with the predictions of the members.

The second stacking version that we explored, dubbed *holdout stacking*, is similar to Kohavi's (1995) holdout accuracy estimation. It differs from the first version, in two ways: the members are not retrained on the entire $L_j$; and each partitioning of $L_j$ into $LL_i$ and $LS_i$ leads to a different president, trained on $LS_i'$, which is then tested on the enhanced $S_j$. Hence, there are $3 \times 10$ presidents in a 10-fold experiment, while in the first version there are only 10. In each case, *WAcc* is averaged over the presidents, and *TCR* is reported as $WErr^b$ over the average *WErr*.

Holdout stacking is likely to be less effective than cross-validation stacking, since its classifiers are trained on smaller sets. Nonetheless, it requires fewer computations, because the members are not retrained. Furthermore, during classification the president consults the same members that were used to prepare its training set. In contrast, in cross-validation stacking the president is tested using members that have received more training than those that prepared its training set. Hence, the model that the president has acquired, which shows when to trust each member, may not apply to the members that the president consults when classifying incoming messages.

## 4  Inducers employed

As already mentioned, we used a Naïve Bayes (NB) and a memory-based learner as members of the committee (Mitchell 1997; Aha, et al. 1991). For the latter, we used TiMBL, an implementation of the *k*-Nearest Neighbor algorithm (Daelemans, et al. 2000).

With NB, the degree of confidence $W_S(\vec{x})$ that $\vec{x}$ is spam is:

$$W_S^{NB}(\vec{x}) = P(spam \mid \vec{x}) =$$

$$= \frac{P(spam) \cdot \prod_{i=1}^{m} P(x_i \mid spam)}{\sum_{k \in \{spam, legit\}} P(k) \cdot \prod_{i=1}^{m} P(x_i \mid k)}$$

NB assumes that $X_1, \ldots, X_m$ are conditionally independent given the category (Duda & Hart, 1973).

With *k*-NN, a distance-weighted method is used, with a voting function analogous to the inverted cube of distance (Dudani 1976). The *k* nearest neighbors $\vec{x}_i$ of $\vec{x}$ are considered:

$$W_S^{k-NN}(\vec{x}) = \frac{\sum_{i=1}^{k} \overline{\delta}(spam, C(\vec{x}_i)) / d(\vec{x}, \vec{x}_i)^3}{\sum_{i=1}^{k} 1 / d(\vec{x}, \vec{x}_i)^3},$$

where $C(\vec{x}_i)$ is the category of neighbor $\vec{x}_i$, $d(\vec{x}_i, \vec{x}_j)$ is the distance between $\vec{x}_i$ and $\vec{x}_j$, and $\overline{\delta}(c_1, c_2) = 1$, if $c_1 = c_2$, and 0 otherwise. This formula weighs the contribution of each neighbor by its distance from the message to be classified, and the result is scaled to [0,1]. The distance is computed by an attribute-weighted function (Wettschereck, et al. 1995), employing *Information Gain* (IG):

$$d(\vec{x}_i, \vec{x}_j) \equiv \sum_{t=1}^{n} IG_t \cdot \delta(x_t^i, x_t^j),$$

where $\vec{x}_i = \langle x_1^i, \ldots, x_m^i \rangle$, $\vec{x}_j = \langle x_1^j, \ldots, x_m^j \rangle$, and $IG_t$ is the *IG* score of $X_t$ (Section 1).

In Tables 1 and 2, we reproduce the best performing configurations of the two learners on Ling-Spam (Androutsopoulos, et al. 2000b; Sakkis, et al. 2001). These configurations were used as members of the committee.

The same memory-based learner was used as the president. However, we experimented with several configurations, varying the neighborhood size (*k*) from 1 to 10, and providing the president with the *m* best word-attributes, as in Section 1, with *m* ranging from 50 to 700 by 50. The same attribute- and distance-weighting schemes were used for the president, as with the ground-level memory-based learner.

| λ | m | SR | SP | TCR |
|---|---|---|---|---|
| 1 | 100 | 82.4% | 99.0% | 5.41 |
| 9 | 100 | 77.6% | 99.5% | 3.82 |

**Table 1:** Best configurations of NB per usage scenario and the corresponding performance.

| λ | k | m | SR | SP | TCR |
|---|---|---|---|---|---|
| 1 | 8 | 600 | 88.6% | 97.4% | 7.18 |
| 9 | 2 | 700 | 81.9% | 98.8% | 3.64 |

**Table 2:** Best configurations of *k*-NN per usage scenario and the corresponding performance.

| λ | true class | only one fails | both fail |
|---|---|---|---|
| 1 | Legitimate | 0.66% | 0.08% |
|   | Spam | 12.27% | 8.52% |
|   | All | 2.59% | 1.49% |
| 9 | Legitimate | 0.33% | 0.08% |
|   | Spam | 19.12% | 10.19% |
|   | All | 3.46% | 1.76% |

**Table 3:** Analysis of the common errors of the best configurations of NB and *k*-NN per scenario (λ) and message class.

Our motivation for combining NB with *k*-NN emerged from preliminary results indicating that the two ground-level learners make rather uncorrelated errors. Table 3 shows the average percentages of messages where only one, or both ground-level classifiers fail, per cost scenario (λ) and message category. The figures are for the configurations of Tables 1 and 2. It can be seen that the common errors are always fewer than the cases where both classifiers fail. Hence, there is much space for improved accuracy, if a president can learn to select the correct member.

## 5 Experimental results

Tables 4 and 5 summarize the performance of the best configurations of the president in our experiments, for each cost scenario. Comparing the *TCR* scores in these tables with the corresponding scores of Tables 1 and 2 shows that stacking improves the performance of the overall filter. From the two stacking versions, cross-validation stacking is slightly better than holdout stacking. It should also be noted that stacking was beneficial for most of the configurations of the president that we tested, i.e. most sub-optimal presidents outperformed the best configurations of the members. This is encouraging, since the optimum configuration is often hard to determine a priori, and may vary from one user to the other.

| λ | k | m | SR | SP | TCR |
|---|---|---|---|---|---|
| 1 | 5 | 100 | 91.7% | 96.5% | 8.44 |
| 9 | 3 | 200 | 84.2% | 98.9% | 3.98 |

**Table 4:** Best configurations of holdout stacking per usage scenario and the corresponding performance.

| λ | k | m | SR | SP | TCR |
|---|---|---|---|---|---|
| 1 | 7 | 300 | 89.6% | 98.7% | 8.60 |
| 9 | 3 | 100 | 84.8% | 98.8% | 4.08 |

**Table 5:** Best configurations of cross-validation stacking per usage scenario and the corresponding performance.

There was one interesting exception in the positive impact of stacking. The 1-NN and 2-NN (*k* = 1, 2) presidents were substantially worse than the other *k*-NN presidents, often performing worse than the ground-level classifiers. We witnessed this behavior in both cost scenarios, and with most values of *m* (number of attributes). In a "postmortem" analysis, we ascertained that most messages misclassified by 1-NN and 2-NN, but not the other presidents, are legitimate, with their nearest neighbor being spam. Therefore, the additional errors of 1-NN and 2-NN, compared to the other presidents, are of the $L \rightarrow S$ type. Interestingly, in most of those cases, both members of the committee classify the instance correctly, as legitimate. This is an indication, that for small values of the parameter *k* the additional two features, i.e., the members' confidence $W_S^1(\vec{x})$ and $W_S^2(\vec{x})$, do not enhance but distort the representation of instances. As a result, the close neighborhood of the unclassified instance is not a legitimate, but a spam e-mail. This behavior of the memory-based classifier is also noted in (Sakkis, et al. 2001). The suggested solution there was to use a larger value for *k*, combined with a strong distance weighting function, such as the one presented in section 4.

## Conclusion

In this paper we adopted a stacked generalization approach to anti-spam filtering, and evaluated its performance. The configuration that we examined combined a memory-based and a Naïve Bayes classifier in a two-member committee, in which another memory-based classifier presided. The classifiers that we chose as members of the committee have been evaluated individually on the same data as in our evaluation, i.e. the Ling-Spam corpus. The results of these earlier studies were used as a basis for comparing the performance of our method.

Our experiments, using two different approaches to stacking and two different misclassification cost scenarios, show that stacking consistently improves the performance of anti-spam filtering. This is explained by the fact that the two members of the committee disagree more often than agreeing in their misclassification errors. Thus, the president is able to improve the overall performance of the filter, by choosing the right member's decision when they disagree.

The results presented here motivate further work in the same direction. In particular, we are interested in combining more classifiers, such as decision trees (Quinlan, 1993) and support vector machines (Drucker, et al. 1999), within the stacking framework. A larger variety of classifiers is expected to lead the president to more informed decisions, resulting in further improvement of the filter's performance. Furthermore, we would like to evaluate other classifiers in the role of the president. Finally, it would be interesting to compare the performance of the stacked generalization approach to other multi-classifier methods, such as boosting (Schapire & Singer, 2000).